\documentclass[preprint,12pt, a4paper]{elsarticle}

\usepackage{amssymb}
\usepackage{hyperref}
\usepackage{xcolor}
\usepackage{listings}
\usepackage{pythonhighlight}

\journal{journal}

\begin{document}
\renewcommand{\labelenumii}{\arabic{enumi}.\arabic{enumii}}

\begin{frontmatter}

\title{KHNNs: hypercomplex neural networks computations via Keras using TensorFlow and PyTorch}

\author[label1]{Agnieszka Niemczynowicz}
\author[label2]{Rados\l{}aw Antoni Kycia}
\address[label1]{Faculty of Mathematics and Computer Science, University of Warmia and Mazury in Olsztyn, S\l{}oneczna 54, Olsztyn, 10-710, Olsztyn, Poland, \texttt{aga.niemczynowicz@gmail.com}}
\address[label2]{Faculty of Computer Science and Telecommunications, Cracow University of Technology, Warszawska 24, Krak\'{o}w, 31-155, Poland, \texttt{kycia.radoslaw@gmail.com}}

\begin{abstract}
Neural networks used in computations with more advanced algebras than real numbers perform better in some applications. However, there is no general framework for constructing hypercomplex neural networks. We propose a library integrated with Keras that can do computations within TensorFlow and PyTorch. It provides Dense and Convolutional 1D, 2D, and 3D layers architectures.
\end{abstract}


\end{frontmatter}

\section*{Metadata}
\label{}

\begin{table}[!h]
\begin{tabular}{|l|p{6.5cm}|p{6.5cm}|}
\hline
\textbf{Nr.} & \textbf{Code metadata description} & \\%\textbf{Please fill in this column} \\
\hline
C1 & Current code version & v1.0.0 \\
\hline
C2 & Permanent link to code/repository used for this code version & TBA \\
\hline
C3  & Permanent link to Reproducible Capsule & None\\
\hline
C4 & Legal Code License   & Apache-2.0 \\
\hline
C5 & Code versioning system used & Git \\
\hline
C6 & Software code languages, tools, and services used & Python 3+ \\
\hline
C7 & Compilation requirements, operating environments \& dependencies &  None \\
\hline
C8 & If available Link to developer documentation/manual & TBA  \\
\hline
C9 & Support email for questions & kycia.radoslaw@gmail.com \\
\hline
\end{tabular}
\caption{Code metadata (mandatory)}
\label{codeMetadata}
\end{table}

\begin{table}[!h]
\begin{tabular}{|l|p{6.5cm}|p{6.5cm}|}
\hline
\textbf{Nr.} & \textbf{(Executable) software metadata description} & \\ %\textbf{Please fill in this column} \\
\hline
S1 & Current software version & v1.0.0 \\
\hline
S2 & Permanent link to executables of this version  & TBA \\
\hline
S3  & Permanent link to Reproducible Capsule & None \\
\hline
S4 & Legal Software License & Apache-2.0 \\
\hline
S5 & Computing platforms/Operating Systems & Python compatible \\
\hline
S6 & Installation requirements \& dependencies & None \\
\hline
S7 & If available, link to user manual - if formally published include a reference to the publication in the reference list & TBA \\
\hline
S8 & Support email for questions & kycia.radoslaw@gmail.com \\
\hline
\end{tabular}
\caption{Software metadata (optional)}
\label{executabelMetadata}
\end{table}

\section{Motivation and significance}

The Artificial Neural Networks (NN) develop in various directions. One of them is the replacement of real numbers computations within the neurons by different hypercomplex algebras like Complex numbers, Quaternions, Clifford algebras, or Octonions. There is a strong suggestion \cite{Marcos1, NiemczynowiczTimeSeries} that such an approach results in NN that has fewer training parameters than the real-numbers approach with similar accuracy.

The Open Source implementation was provided for some four-dimensional hypercomplex algebras in \cite{Marcos1}. This implementation requires the computation of an algebra multiplication matrix to include new algebras. It also works only for four-dimensional data. In \cite{NiemczynowiczKyciaTheory}, the theoretical aspects of generalization for all possible algebras, including hypercomplex ones, were given. In this paper we describe an example implementation.

There are some alternative approaches, e.g., \cite{ParametrizedHypercomplexNN} that presents Parametrized Hypercomplex Neural Networks, which adjust hyperalgebra to the data. However the implementation is limited to PyTorch. The hyperalgebra in this approach cannot be treated as a hyperparameters, and moreover, the focus is only on hypercomplex algebras and not general algebraic structures. The TensorFlow implementation is still missing.

The standard industrial and research framework for constructing feed-forward NN is the Keras high-level interface that uses TensorFlow \cite{Tensorflow} or PyTorch \cite{Pytroch} as the backend. The library described here extends this common architecture for arbitrary (hypercomplex) algebras computations capabilities.

The KHNN library provides Dense and Convolutional 1D, 2D, and 3D layers that can be included in any feed-forward architecture. Therefore, there are unlimited ways to use this library in research experiments, data analysis, and industrial applications.

\section{Software description}

The library is based on Keras and has two branches: TensorFlow and PyTorch. This means there are Dense and Convolutional layers that use internal TensorFlow, and PyTorch computations.

The library has predefined algebras like Complex numbers, Quaternions, Klein four-group, Clifford algebra (2,0), Clifford algebra (1,1), Bicomplex numbers, Tessarines, and Octionions. However it has an easy way to implement arbitrary algebra computations.

The workflow with the library is standard and is as follows:
\begin{enumerate}
 \item {Import algebra module and select or define algebra to work with.}
 \item {Import desired layers}
 \item {Construct neural network from the layers}
 \item {Train and tune NN}
 \item {Make predictions}
\end{enumerate}

\subsection{Software architecture}

The KHNN is a divided into three logical parts:
\begin{itemize}
 \item{\texttt{Algebra} module: contains the \texttt{StructureConstants} class that allows to define multiplication of an algebra; contains also predefined multiplication tables for various algebras: \texttt{Complex}, \texttt{Quaternions}, \texttt{Klein4}, \texttt{Cl20} - Clifford (2,0) algebra , \texttt{Coquaternions}, \texttt{Cl11}- Clifford (1,1), \texttt{Bicomplex}, \texttt{Tessarines}, \texttt{Octonions};}
\item{\texttt{Keras + TensorFlow} part contains:
\begin{itemize}
 \item {\texttt{Hyperdense} module that contains \texttt{HyperDense} class realizing hypercomplex Dense layer;}
 \item {\texttt{Convolutional} module that contains \texttt{HyperConv1D}, \texttt{HyperConv2D}, \texttt{HyperConv3D};}
\end{itemize}
}
\item {\texttt{Keras + PyTorch} part that contains:
\begin{itemize}
 \item {\texttt{HyperdenseTorch} module that contains \texttt{HyperDenseTorch} class realizing hypercomplex Dense layer;}
\end{itemize}
}
\end{itemize}

\subsection{Software functionalities}

The software have two types of functionality: Algebra manipulations and NN construction.

The algebra computations are realized by \texttt{Algebra} module. The basic class is \texttt{StructureConstants}, which realizes multiplication within the algebra. We summarize the theory briefly from \cite{NiemczynowiczKyciaTheory}. Assume that the algebra has a base $\{e_{i}\}_{i=0}^{n-1}$, where $n$ is the dimension of algebra. One assumes that $e_{0}$ is the multiplication unit. Then the multiplication is defined by the tensor $e_{i}\cdot e_{j}=A_{ijk}e_{k}$. An example of a multiplication table is given in (\ref{Eq_multiplicationTable}).
\begin{equation}
 \begin{array}{c|c}
   \cdot & e_{j} \\ \hline
   e_{i} & A_{ijk}e_{k}
 \end{array}
 \label{Eq_multiplicationTable}
\end{equation}
The way of defining a multiplication matrix is to define the dictionary where the entry is $(i,j):(k,A_{ijk})$.

As a simple example define complex numbers (already defined in library) given by the multiplication table (\ref{Eq_multiplicationTableComplex}).
\begin{equation}
 \begin{array}{c|c|c}
   \cdot & e_{0} = 1  & e_{1} = i \\ \hline
   e_{0}=1 & e_{0} & e_{1} \\ \hline
   e_{1}=i & e_{1} & -e_{0}
 \end{array}
 \label{Eq_multiplicationTableComplex}
\end{equation}
This gives
\begin{python}
#Define dictionary for complex numbers (implicitly assumed that e_0 is the unit of multiplication)
Complex_dict = {(1,1):(0,-1)}
#Define multiplication constants
Complex = StructureConstants(Complex_dict )
#Example operations:
## 1 x 1
Complex.Mult(np.array([1,0]), np.array([1,0]))  # gives 1
## i x i
Complex.Mult(np.array([0,1]), np.array([0,1])   # gives -1
#Get multiplication tensor
Complex.getA()
\end{python}
%%%%

The second type is to define neural networks, which will be presented in the following subsection.

\section{Illustrative examples}

We give some elementary examples of applications of the KHNN library. The first example will be related to \texttt{HyperDense} layer for quaternions.

The example for TensorFlow is presented below.
\begin{python}
import numpy as np
from keras.models import Sequential
from keras.layers import Dense, Activation

from Hyperdense import HyperDense

#Preparation of data:
x_train = np.array([[1,0, 0, 0], [0, 1, 0, 0], [0, 0, 1, 0],  [0, 0, 0, 1]], dtype = np.dtype(float))
y_train = np.array([[0], [1], [1],  [0]])

#Define model:
model = Sequential()
num_neurons = 4
model.add(HyperDense(num_neurons))
#model.add(Dense(num_neurons))  #real numbers alternative for comparision
model.add(Activation('tanh'))
model.add(Dense(1))
model.add(Activation('sigmoid'))

#Setup learning
opt = tf.keras.optimizers.legacy.Adam()
model.compile(loss='binary_crossentropy', optimizer=opt, metrics=['accuracy'])

#Train model
model.fit(x_train, y_train, epochs=500, verbose=0)

#Make prediction
y_predict = model.predict(x_train, verbose=0)
y_predict_quantized = np.round(y_predict).astype(int)
\end{python}

The same code using PyTorch implementation:
\begin{python}
import torch
import torch.nn as nn
from collections import OrderedDict
import matplotlib.pylab as plt

from HyperdenseTorch import HyperDenseTorch

#Preparation of data:
x_train = torch.Tensor(np.array([[1, 0, 0, 0], [0, 1, 0, 0], [0, 0, 1, 0],  [0, 0, 0, 1]], dtype = np.dtype(float))).to(torch.float)
y_train = torch.Tensor(np.array([[0], [1], [1],  [0]])[:,0]).to(torch.float)

#Define model:
model = nn.Sequential(OrderedDict([
    ("HyperDense", HyperDenseTorch(10, (4,), activation = torch.tanh )),
    ("Dense", nn.Linear(40,1)),
    ('Sigmoid', nn.Sigmoid())
        ]))

#Setup learning
loss_fn = nn.BCELoss()
optimizer = torch.optim.SGD(model.parameters(), lr=0.015)
torch.manual_seed(1)

num_epoch = 200

loss_hist_train = [0]*num_epoch
accuracy_hist_train = [0]*num_epoch
loss_hist_train = [0]*num_epoch

#training loop
for epoch in range(num_epoch):
    pred = model(x_train)[:,0]
    #pred = model(x_train)
    #print("epoch = ", epoch)
    #print("pred = ", pred)
    #print("y_train = ", y_train)
    loss = loss_fn(pred, y_train)
    loss.backward()
    optimizer.step()
    optimizer.zero_grad()
    loss_hist_train[epoch] += loss.item()
    is_correct  = ((pred >= 0.5).float() == y_train).float()
    accuracy_hist_train[epoch] += is_correct.mean()

#Generate summary
pred = model(x_train)[:,0]
print("predicted = ", pred)
print("predicted (rounded) = ", pred.round())
print("expected = ", y_train)

plt.plot(loss_hist_train, label = "loss")
plt.plot(accuracy_hist_train, label = "acuracy")
plt.legend()
plt.show()
\end{python}

\begin{figure}
\centering
 \includegraphics[width = 0.7\textwidth]{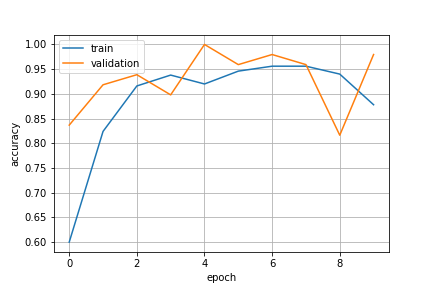}
 \caption{Accuracy for training and validation data during fitting the model.}
 \label{Fig_accuracy}
\end{figure}

\begin{figure}
\centering
 \includegraphics[width = 0.7\textwidth]{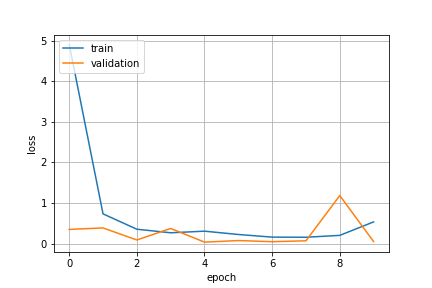}
 \caption{Loss function for training and validation data during fitting the model.}
 \label{Fig_loss}
\end{figure}

The final example presents the usage of 2-dimensional hypercomplex convolutional NN in image classification using TensorFlow. We select the blood images with and without malaria from \cite{MalariaDataset}. Since the color encoding is RGB, we adjusted the color information to ARGB by adding channel Alpha set to zero. Thanks to this, we can encode color data in four-dimensional algebra\footnote{The alpha channel is associated with a unit of the algebra. It is typical to associate with the algebra unit some distinguished data axes.}. The following code do the analysis.
\begin{python}
import pandas as pd
import matplotlib.pyplot as plt
import numpy as np

#Load data
import tensorflow_datasets as tfds
tfds.list_builders()
ds = tfds.load('malaria', split='train', shuffle_files=True)
#Select first 700 records
X = []
Y = []
i=0
for example in ds:
    image = example["image"]
    label = example["label"]
    X.append(image)
    Y.append(label)
    i += 1
    if i > 700:
        break
X = list(map(lambda image: tf.image.resize(image, (100, 100)),X))
X = np.array(X)
Y = np.array(Y)
import tensorflow_io as tfio
X4 = tfio.experimental.color.rgb_to_rgba(X)
#Do abgr
X4 = tf.reverse(X4,[-1])
#Quantize labels
idxY = np.logical_or(Y==0, Y == 1)
X_data = X4[idxY]
Y_data = Y[idxY]
#Do deep learning

import Algebra
from Convolutional import HyperConv2D
from Hyperdense import HyperDense

import numpy as np
from keras.models import Sequential
from keras.layers import Dense, Activation, GlobalMaxPooling2D, Dropout
from keras.layers import Dense, Activation, MaxPooling2D, Dropout, Flatten

#Split data:
x_train = tf.cast(X_data, tf.float32)[:500]
y_train = np.asarray(Y_data).astype('int').reshape((-1,1))[:500]
x_validate = tf.cast(X_data, tf.float32)[501:550]
y_validate = np.asarray(Y_data).astype('int').reshape((-1,1))[501:550]
x_test = tf.cast(X_data, tf.float32)[551:]
y_test = np.asarray(Y_data).astype('int').reshape((-1,1))[551:]

#Create model:
num_neurons = 100
hidden_dims = 20
model = Sequential()
model.add(HyperConv2D(num_neurons, (3,3), algebra=Algebra.Quaternions))
model.add(GlobalMaxPooling2D())
model.add(Dense(1))
model.add(Activation('sigmoid'))

model.predict(x_train, verbose=0)
model.summary()
opt = tf.keras.optimizers.legacy.Adam()
model.compile(loss='binary_crossentropy', optimizer=opt, metrics=['accuracy'])
#Do learning
history = model.fit(x_train, y_train, validation_data=(x_validate, y_validate), epochs=10, verbose=1)

print("Evaluate on test data")
results = model.evaluate(x_test, y_test, batch_size=10)
print("evaluation = ", results)

plt.plot(history.history['accuracy'])
plt.plot(history.history['val_accuracy'])
plt.ylabel('accuracy')
plt.xlabel('epoch')
plt.legend(['train', 'val'], loc='upper left')
plt.grid()
plt.show()

plt.plot(history.history['loss'])
plt.plot(history.history['val_loss'])
plt.ylabel('loss')
plt.xlabel('epoch')
plt.legend(['train', 'val'], loc='upper left')
plt.grid()
plt.show()
\end{python}
Which produces Figs. \ref{Fig_accuracy} and \ref{Fig_loss}.

\section{Impact}
Currently, applications of hypercomplex algebras in neural networks and usage in various disciplines are beginning. Usually, research focuses only on a small range of algebras due to a case-by-case approach to implementation. The presented library makes significant progress in the field by providing a general framework for the broad application of such NN.

When data naturally lump into tuples, one can always try to find an algebra of data lump that encodes
a single piece of data in its representative, and then, process it naturally as a whole.

Since NN has various applications, the usefulness of this library is immense.

\section{Conclusions}
KHNN is a small, versatile library that updates the Keras interface (both in TensorFlow and PyTorch) for hypercomplex Dense and Convolutional layers. It can be extended easily for any algebra. Thanks to this, it can become an essential research tool for hypercomplex neural networks and applications.

\section*{Acknowledgements}
This paper has been supported by the Polish National Agency for Academic Exchange Strategic Partnership Programme under Grant No. BPI/PST/2021/1/00031 (nawa.gov.pl).

\end{document}